%% file: main.tex
\documentclass[letterpaper]{article} 

\usepackage[preprint]{aaai2027}  
\usepackage[hyphens]{url}  
\usepackage{graphicx} 
\usepackage{natbib}  
\usepackage{caption} 
\usepackage{amsmath}
\usepackage{amssymb}
\usepackage{booktabs}
\usepackage{tabularx}
\usepackage{array}
\usepackage{multirow}

\newcommand{\eg}{e.g.}

\title{InspectorGPT: A Comparative Reasoning Enhanced VLM 
for Comprehensive Industrial Anomaly Detection}

\author{
    Weifei Chen\textsuperscript{\rm 1},
    Honghao Zhang\textsuperscript{\rm 1},
    Zhiyuan You\textsuperscript{\rm 2},
    Xinyi Le\textsuperscript{\rm 1}
}
\affiliations{
    \textsuperscript{\rm 1}Shanghai Jiao Tong University\\
    \textsuperscript{\rm 2}The Chinese University of Hong Kong
}

\begin{document}

\maketitle

\begin{abstract}
  Industrial anomaly detection is a critical component of modern manufacturing. Most traditional unsupervised methods rely on modelling normal feature distributions, inherently limiting generalization to unknown categories. To improve generalizability, some recent methods incorporate vision-language models (VLMs) for zero-shot detection via text prompts. However, we observe that reasoning-oriented post-training can cause anomaly discrimination to collapse, with some fine-tuned models performing worse than their base VLMs. Existing methods also provide only textual decisions or coarse boxes, without pixel-level segmentation. A more explicit detection principle comes from human inspection: anomalies are identified by comparing a query image with a defect-free reference. Inspired by this, we propose InspectorGPT, a VLM framework centered on comparative reasoning. Given a normal reference and a query image, InspectorGPT compares them to identify discrepancies and perform multiple inspection tasks with detailed reasoning. We internalize this capability through Chain-of-Thought (CoT) fine-tuning and Group Relative Policy Optimization (GRPO) with tailored, verifiable rewards. We further introduce InspectorGPT-Seg for pixel-level anomaly masks. Segmentation supervision improves anomaly discrimination but weakens semantic reasoning, while joint training fails to balance them. We therefore train the two branches separately and combine them through task-vector fusion. Extensive experiments demonstrate superior multi-dimensional performance and generalization to unseen benchmarks, validating comparative reasoning for comprehensive industrial inspection.
\end{abstract}

\input{content/introduction}
\input{content/related_work}
\input{content/method}
\input{content/experiment}
\input{content/conclusion}

\small
\bibliography{main}

\end{document}

%% file: content/introduction.tex
\section{Introduction}
\label{sec:introduction}

\begin{figure*}[t]
    \centering
    \includegraphics[width=0.90\textwidth]{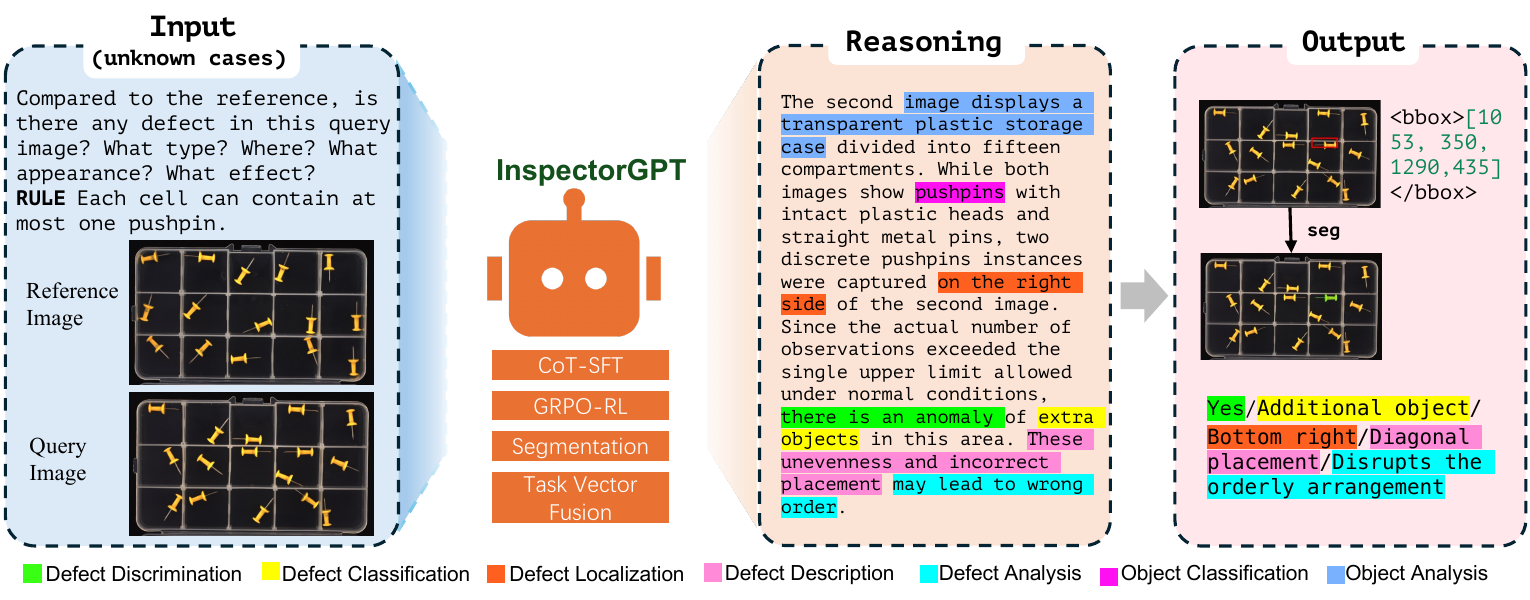}
    \caption{Multi-dimensional comparative reasoning in InspectorGPT.
    Given a normal reference image and a query image, our model performs structured comparative reasoning and generates a Chain-of-Thought covering diverse inspection dimensions (color-coded).}
    \label{fig:introduction}
\end{figure*}

Industrial Anomaly Detection (IAD) is a core component of manufacturing quality control, aiming to identify and localize image regions that deviate from normal patterns. As industrial production moves towards higher variety and faster iteration, anomaly detectors are increasingly required to transfer across categories. Traditional pure-vision methods~\citep{defard2021padim,yu2021fastflow,you2022unified} learn a normal distribution (\eg, the ``one cell one pin'' distribution in Figure~\ref{fig:introduction}) from defect-free samples and derive anomaly scores from feature distances~\citep{roth2022towards,cohen2020sub}. However, when product types change, normal prototypes for new types are unavailable without retraining. Their binary anomaly scores also lack semantic descriptions for defect identification and interpretable evidence.

To enhance generalization and interpretability, recent studies have adopted Vision-Language Models (VLMs) as backbone architectures for anomaly detection. CLIP-based approaches~\citep{jeong2023winclip,cao2024adaclip,qu2024vcp} enable zero-shot detection via text prompts, but rely on accurate class definitions or hand-crafted priors. More recent methods~\citep{gu2024anomalygpt,li2025triad,xu2025towards} therefore turn to multimodal large language models (MLLMs), which produce fluent and interpretable explanations. To further align these models with anomaly inspection, several approaches post-train them on anomaly data using CoT supervision and reinforcement learning~\citep{li2025iadr1,chao2025anomalyr1,liao2026adfm,kang2026judo}. However, two critical issues remain. First, anomaly discrimination, the most fundamental capability of detection, can collapse after reasoning-oriented post-training: some fine-tuned models score even lower than their own base VLMs~\citep{kang2026judo}. This suggests that the model may learn dataset-dependent response patterns rather than a transferable detection principle. Second, localization outputs are often limited to text decisions or coarse bounding boxes. This granularity is insufficient for pixel-level defect delineation and quantitative assessment.

We hypothesize that the discrimination collapse occurs because the model is taught to imitate answers rather than to ground its decisions in concrete visual discrepancies. Human experts instead follow a simple principle: given a defect-free reference, they judge each region of the query image through comparison. Each judgment is thus anchored to an observable discrepancy. Such comparative reasoning provides an explicit detection principle that can transfer beyond category-specific appearances.

Inspired by human perception, we propose \textbf{InspectorGPT}, a comparative-reasoning-enhanced framework.
Our InspectorGPT elevates a general-purpose VLM into an industrial expert by teaching it structured comparative reasoning.
Specifically, it works with three processes.
First, InspectorGPT takes a normal reference image as a template and compares the query image with the normal template (Figure~\ref{fig:introduction}, left).
Then, as illustrated in the middle of Figure~\ref{fig:introduction}, it performs semantic-level comparisons and generates a structured Chain-of-Thought covering diverse inspection dimensions.
Finally, based on the detailed reasoning results, InspectorGPT performs multiple tasks including anomaly discrimination, detection, classification, segmentation, and so on (Figure~\ref{fig:introduction}, right).
To empower InspectorGPT with these capabilities, we leverage Chain-of-Thought (CoT) fine-tuning alongside Group Relative Policy Optimization (GRPO) with tailored rewards.
We introduce InspectorGPT-Seg, integrating a segmentation module to enhance its localization and segmentation abilities.

Interestingly, we observe that dense segmentation supervision substantially improves anomaly discrimination, but fails to fully preserve the semantic reasoning capability. Directly optimizing the reasoning and segmentation objectives in a single model also fails to balance the two capabilities. We therefore train the reasoning and segmentation branches separately from a shared SFT checkpoint and assemble them through task-vector fusion. This design preserves most of the semantic reasoning capability while recovering anomaly discrimination.

By anchoring its decisions to reference-conditioned discrepancies rather than a category-specific fixed normal prototype, InspectorGPT can generalize to unseen benchmarks without target-domain adaptation. The main contributions of this work are summarized as follows:
\begin{itemize}
    \item We propose InspectorGPT, a multimodal framework that trains a VLM to facilitate comparative reasoning for industrial inspection. Training combines CoT-style supervised fine-tuning and GRPO with task-specific rewards, prompting the model to internalize generalizable detection principles. The model supports interpretable outputs and strong transfer to novel categories without retraining.
    \item We identify the discrimination collapse of reasoning-oriented post-training and address it with a staged training strategy combined with task vector fusion. We further introduce InspectorGPT-Seg, which couples a lightweight segmentation decoder to produce pixel-level anomaly masks instead of coarse boundaries.

    \item Extensive experiments demonstrate superior multi-dimensional performance and generalization to unseen benchmarks, outperforming existing commercial, open-source, and AD-specific models. These results validate comparative reasoning as a robust paradigm for comprehensive industrial inspection.
\end{itemize}

%% file: content/related_work.tex
\section{Related Work}
\label{sec:related_work}

\noindent\textbf{Traditional Industrial Anomaly Detection.}
Classical methods learn a model of normality from defect-free samples and flag deviations from it as anomalies. They broadly fall into two paradigms: embedding-based methods that compare test features against stored normal representations via memory banks, student--teacher discrepancy, or one-class boundaries~\citep{roth2022towards, defard2021padim, bergmann2020uninformed, salehi2021multiresolution}, and reconstruction-based methods that treat large reconstruction errors as anomaly signals~\citep{park2020learning,zavrtanik2021reconstruction}. Despite strong benchmark accuracy, both paradigms model category-specific normality rather than transferable detection principles. Thus, they degrade on categories unseen during training without additional fine-tuning. Moreover, they output only anomaly scores or binary decisions, offering no semantic description or interpretable evidence for downstream decisions.

\noindent\textbf{Vision-Language Models for Anomaly Detection.}
Large pre-trained vision-language models bring the possibility of general-purpose anomaly detection. Recent work adopts multimodal large language models (MLLMs) that follow instructions and produce interpretable reasoning, evaluated by benchmarks such as MMAD~\citep{jiang2024mmad}. A representative early design pairs an expert perception module with a frozen LLM~\citep{gu2024anomalygpt}, where the LLM mainly explains the module's outputs. To make reasoning end-to-end and better aligned, subsequent methods post-train the MLLM itself on anomaly data, typically with chain-of-thought supervision and GRPO-style reinforcement learning~\citep{shao2024deepseekmath,li2025iadr1,chao2025anomalyr1,liao2026adfm,kang2026judo}. However, such reasoning-oriented post-training exposes two limitations: the anomaly discrimination accuracy of a post-trained model often drops below that of its own base VLM~\citep{kang2026judo}, and the outputs remain textual answers or coarse boxes rather than pixel-level localization. Dense masks can be obtained by adapting foundation models like SAM~\citep{kirillov2023segment} for zero-shot anomaly segmentation~\citep{cao2023saa, li2025clipsam, zhou2024anomalyclip}, and a recent attempt derives anomaly maps by aggregating the attention of reasoning tokens~\citep{jin2026readl}. Nevertheless, pixel-level segmentation is still rarely supported among VLM-based detectors.

\noindent\textbf{Task Vector Fusion.}
Fine-tuning can be viewed as a displacement in weight space. Task arithmetic~\citep{ilharco2023task} shows that such displacements, termed task vectors, compose approximately linearly: adding them equips a single model with multiple abilities without joint training. Follow-up work improves the composition by averaging multiple fine-tuned models~\citep{wortsman2022soups} or resolving parameter interference during merging~\citep{yadav2023ties}. These techniques offer a training-free way to balance abilities that conflict under joint optimization. Such fusion allows VLM-based detection models to balance capabilities across different task dimensions.

%% file: content/method.tex
\section{Method}
\label{sec:method}

\begin{figure*}[t]
    \centering
    \includegraphics[width=0.85\textwidth]{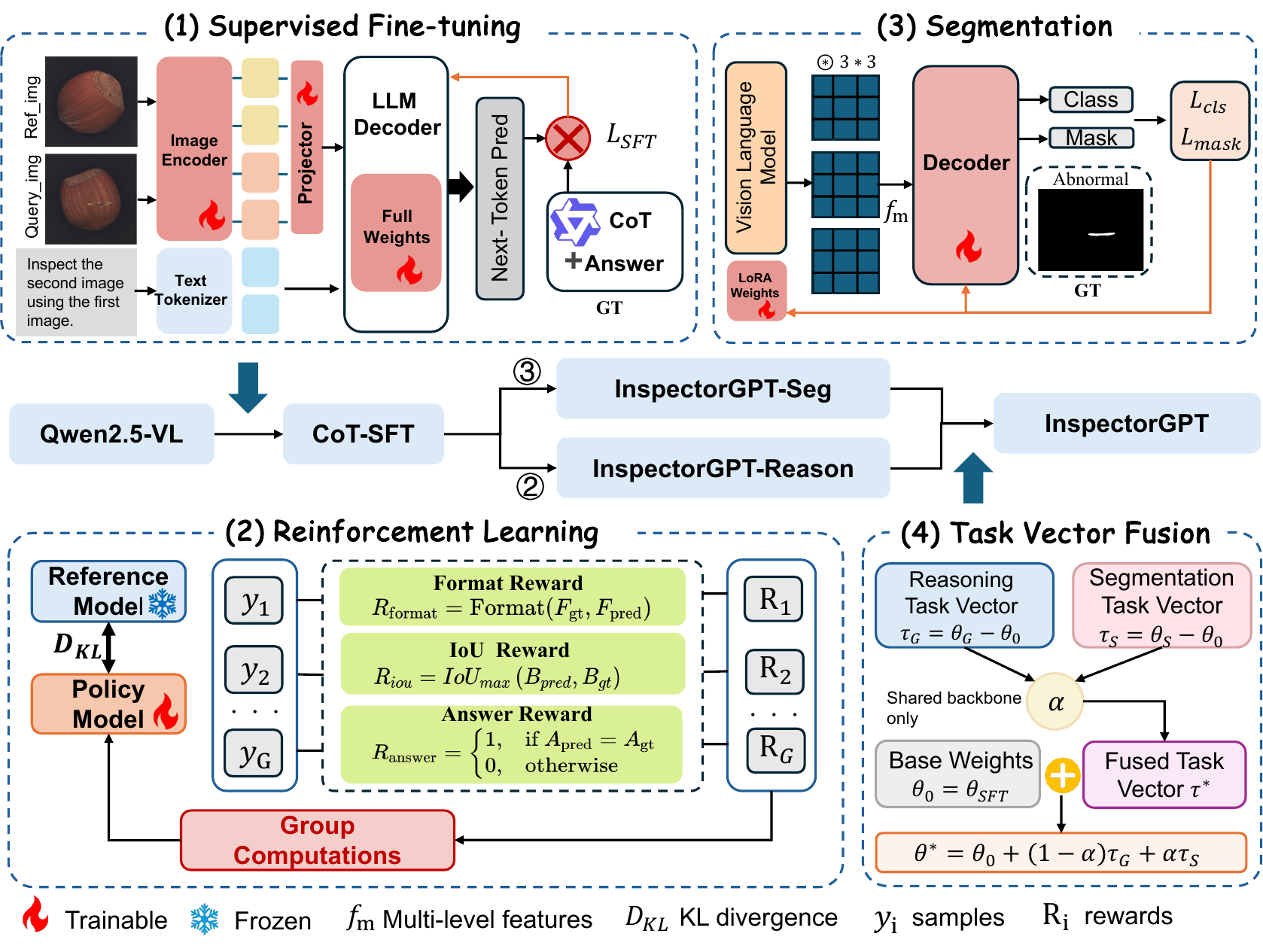}
    \caption{Overview of the InspectorGPT framework. Our approach consists of four stages: (1) supervised fine-tuning on a comparative CoT formulation; (2) Group Relative Policy Optimization (GRPO) using format, IoU, and answer rewards; (3) a parallel segmentation branch for pixel-level anomaly grounding; and (4) task-vector fusion.}
    \label{fig:method}
\end{figure*}

Our framework contains four stages, as illustrated in Figure~\ref{fig:method}: (1) Chain-of-Thought Supervised Fine-Tuning, (2) Reinforcement Learning via Group Relative Policy Optimization, (3) Segmentation Grounding, and (4) Task-Vector Fusion. After CoT-SFT, the reinforcement learning and segmentation branches are optimized in parallel from the same SFT checkpoint. The objective is to learn a transferable comparative reasoning principle and to produce grounded outputs.

\subsection{Stage 1: Chain-of-Thought Supervised Fine-Tuning}
\label{subsec:sft}

This stage initializes the model with an explicit comparative reasoning pattern. The model receives a reference image, a query image, and an inspection prompt. It is trained to generate a structured Chain-of-Thought.

\noindent\textbf{CoT data construction.}
To efficiently scale high-quality reasoning trajectories, we employ Qwen3-Plus as an expert annotator. For each training sample, it receives the defect-free reference $I_r$, the query $I_q$, and the ground-truth mask $I_m$. The mask is used only as a spatial hint to encourage grounded discrepancy descriptions. Crafted prompts constrain the generated trajectories to follow a three-step \texttt{[Align-Contrast-Diagnose]} paradigm. After manual verification, these trajectories form our labeled CoT data.

\noindent\textbf{Optimization.}
We perform full-parameter supervised fine-tuning on Qwen2.5-VL~\citep{bai2025qwen25vltechnicalreport} to enforce the comparative format across both vision and language components. The SFT objective is the standard autoregressive loss:
\begin{equation}
\mathcal{L}_{\text{SFT}} = - \sum_{t=1}^{T} \log \pi_\theta(y_t \mid x, \mathbf{y}_{<t}),
\label{eq:sft}
\end{equation}
where $\pi_\theta$ is the VLM policy. The resulting model is denoted as $\pi_{\text{ref}}$, serving as the frozen reference policy for the subsequent reinforcement learning phase. Its shared VLM weights are denoted by $\theta_0=\theta_{\text{SFT}}$, which initialize both parallel branches.

\subsection{Stage 2: GRPO Alignment with Verifiable Rewards}
\label{subsec:grpo}

SFT provides a valid reasoning structure but does not directly optimize inspection metrics or prevent unsupported answers. We therefore apply Group Relative Policy Optimization (GRPO)~\citep{shao2024deepseekmath} to align generation with task objectives.

\noindent\textbf{GRPO Mechanism.}
For each input $x$, we sample a group of $G$ completions $\{y_i\}_{i=1}^{G}$ from the active policy $\pi_\theta$. Each completion receives a scalar reward $R_i$. We compute group-normalized advantages and update $\pi_\theta$ with a KL constraint to $\pi_{\text{ref}}$:
\begin{equation}
\mathcal{L}_{\text{GRPO}} =
- \frac{1}{G} \sum_{i=1}^{G}
\Big[
A_i \log \pi_\theta(y_i \mid x)
- \beta \, \mathbb{D}_{\text{KL}}(\pi_\theta \,\|\, \pi_{\text{ref}})
\Big].
\label{eq:grpo}
\end{equation}

\noindent\textbf{Designed Reward Engine.}
The core of our reinforcement learning stage lies in the customized reward formulation. Using strict tag-based parsing rules, each trajectory $y_i$ is decoupled into a predicted bounding box set $B_{\text{pred}}$ and a diagnostic answer $A_{\text{pred}}$. We evaluate these outputs across three dimensions:

\begin{itemize}
    \item \textbf{Format Reward.} We explicitly reward outputs that maintain parsable, structured XML reasoning tags:
    \begin{equation}
    \begin{split}
    R_{\text{format}}(y_i) = {}& \mathbb{I}[\texttt{<answer>}] + \mathbb{I}[\texttt{<bbox>}] \\
    & + \mathbb{I}[\texttt{<think>}_{\text{non-empty}}],
    \end{split}
    \label{eq:rfmt}
    \end{equation}
    where $\mathbb{I}[\cdot]$ denotes the indicator function.

    \item \textbf{Conditional IoU Reward.} Industrial datasets frequently contain normal samples where the ground-truth box set $B_{\text{gt}}$ is empty ($\varnothing$). To handle this asymmetry, we define
    \begin{equation}
    R_{\text{iou}}(y_i) =
    \begin{cases}
    1, & B_{\text{gt}} = \varnothing,\, B_{\text{pred}} = \varnothing \\
    \max\limits_{b,\, g} \mathrm{IoU}(b, g), & B_{\text{gt}} \neq \varnothing,\, B_{\text{pred}} \neq \varnothing \\
    0, & \text{otherwise},
    \end{cases}
    \label{eq:riou}
    \end{equation}
    where the maximum is taken over $b \in B_{\text{pred}}$ and $g \in B_{\text{gt}}$.

    \item \textbf{Answer Reward with Verifiability Gate.} To suppress ungrounded reward hacking, we gate the binary answer match with a spatial verifiability constraint $\mathbb{I}_{\text{gate}}$:
    \begin{equation}
    R_{\text{answer}}(y_i) = \mathbb{I}[A_{\text{pred}} = A_{\text{gt}}] \cdot \mathbb{I}_{\text{gate}},
    \label{eq:rgate}
    \end{equation}
    where $\mathbb{I}_{\text{gate}} = 0$ if the model predicts an anomaly without a corresponding bounding box or predicts a box for a normal sample. Otherwise, $\mathbb{I}_{\text{gate}} = 1$.
\end{itemize}

Finally, the total scalar reward is aggregated via non-negative weights:
\begin{equation}
R_i = \lambda_{\text{fmt}} R_{\text{format}}(y_i) + \lambda_{\text{iou}} R_{\text{iou}}(y_i) + \lambda_{\text{ans}} R_{\text{answer}}(y_i).
\label{eq:rtotal}
\end{equation}
The resulting reasoning model is denoted as \textbf{InspectorGPT-Reason}, with shared backbone weights $\theta_G$.

\subsection{Stage 3: Segmentation Grounding}
\label{subsec:decoder}

We introduce a segmentation branch to upgrade coarse bounding boxes into pixel-level masks, while keeping the VLM output interpretable. This branch is initialized from $\theta_0$ in parallel with the GRPO branch.

\noindent\textbf{Spatial Context Adapters.}
Rather than standard linear projections, we extract multi-scale feature grids from the VLM backbone and process them through $3 \times 3$ Convolutional Spatial Context Adapters. This preserves high-frequency spatial textures and reduces fragmented masks.

\noindent\textbf{Dense Defect Localization Loss.}
As illustrated in Figure~\ref{fig:method}(3), the adapted multi-scale features are fed into a decoder adapted from Mask2Former~\citep{cheng2022masked}. It outputs a categorical anomaly probability $P_{\text{cls}}$ and a dense anomaly mask $P_{\text{mask}}$. The global loss is defined as $\mathcal{L}_{\text{total}} = \lambda_{\text{cls}}\mathcal{L}_{\text{cls}} + \mathcal{L}_{\text{mask}}$, where
\begin{equation}
\mathcal{L}_{\text{mask}} = \lambda_{\text{bce}} \mathcal{L}_{\text{BCE}} + \lambda_{\text{dice}} \mathcal{L}_{\text{Dice}} + \lambda_{\text{tv}} \mathcal{L}_{\text{TV}} + \lambda_{\text{const}} \mathcal{L}_{\text{const}}.
\label{eq:mask_loss}
\end{equation}
Here, $\mathcal{L}_{\text{BCE}}$ and $\mathcal{L}_{\text{Dice}}$ ensure pixel alignment under severe class imbalance. The total-variation term $\mathcal{L}_{\text{TV}}=\|\nabla P_{\text{mask}}\|_1$ encourages spatial continuity~\citep{rudin1992nonlinear}. The consistency term $\mathcal{L}_{\text{const}}=(P_{\text{cls}}-\max P_{\text{mask}})^2$ aligns global classification with local activation.

\noindent\textbf{Progressive LoRA Co-training.}
After training the decoder with a frozen VLM, we inject LoRA modules~\citep{hu2022lora} into the attention and MLP projection layers. We then jointly optimize the LoRA parameters and decoder weights using a reduced learning rate. This provides a direct gradient pathway from the pixel-level objective to the VLM. After merging the learned LoRA updates, we obtain the shared backbone weights $\theta_S$ and retain the decoder $\Phi_S$. Together they form \textbf{InspectorGPT-Seg}.

\subsection{Stage 4: Task-Vector Fusion}
\label{subsec:fusion}

Task arithmetic represents a learned capability as the parameter displacement between a fine-tuned model and its initialization~\citep{ilharco2023task}. Task vectors derived from a common initialization can be linearly combined without additional training. However, we observe that joint optimization fails to balance anomaly discrimination and reasoning, as improving one often compromises the other. We therefore train the two branches separately from the common SFT weights $\theta_0$ and combine them post hoc. GRPO yields the InspectorGPT-Reason weights $\theta_G$, while segmentation training yields the InspectorGPT-Seg weights $\theta_S$. Their task vectors are defined as
\begin{equation}
\begin{gathered}
\tau_G = \theta_G - \theta_0, \qquad \tau_S = \theta_S - \theta_0, \\
\tau^* = (1-\alpha)\tau_G + \alpha\tau_S, \qquad \theta^* = \theta_0 + \tau^*.
\end{gathered}
\label{eq:fusion}
\end{equation}
Only the shared backbone weights are fused. The segmentation decoder $\Phi_S$ is retained separately. We use $\alpha=0.3$ for the final InspectorGPT.

%% file: content/experiment.tex
\section{Experiments}
\label{sec:experiments}
\subsection{Experimental Setup}
\label{subsec:setup}

\noindent\textbf{Datasets \& Protocols.} We evaluate InspectorGPT on MMAD~\citep{jiang2024mmad}, which aggregates four diverse benchmarks: MVTec-AD~\citep{bergmann2019mvtec}, VisA~\citep{zou2022spot}, MVTec-LOCO~\citep{bergmann2022beyond}, and GoodsAD~\citep{zhang2024pku}. Performance is measured over seven VQA dimensions: Anomaly Discrimination, Defect Classification, Defect Localization, Defect Description, Defect Analysis, Object Classification, and Object Analysis. We train on a 20\% stratified subset of MMAD with uniform coverage across all 38 categories and a 1:1 normal-to-anomalous ratio, and test on the remaining 80\%, following the split protocol of recent fine-tuned methods~\citep{zhao2025omniaddetectunderstandindustrial,liao2026adfm}. To further assess generalization, the same trained model is directly evaluated on four benchmarks disjoint from MMAD: DAGM~\citep{wieler2007dagm}, DTD-Synthetic (DTD)~\citep{aota2023zero}, SDD~\citep{tabernik2020sdd}, and MPDD~\citep{jezek2021mpdd}. For reference-image selection, we randomly sample from the reference images originally provided by MMAD.

\noindent\textbf{Compared Methods.}
We compare against three types of baselines. Proprietary general-purpose VLMs include GPT-4o, GPT-4o-mini, Gemini 2.5 Pro, and Qwen3-VL-Plus; open-source VLMs include Qwen2.5-VL (7B/72B)~\citep{bai2025qwen25vltechnicalreport}, LLaVA variants~\citep{li2024llava,lif2024llava}, and the InternVL family~\citep{chen2024internvl,wang2025internvl3}. All general-purpose models are tested under the comparative protocol, receiving the query image together with a defect-free reference. We further benchmark against IAD-specialized fine-tuned MLLMs under the same MMAD formulation: AnomalyR1~\citep{chao2025anomalyr1}, IAD-R1~\citep{li2025iadr1}, OmniAD~\citep{zhao2025omniaddetectunderstandindustrial}, EMIT~\citep{guan2025emit}, AD-FM~\citep{liao2026adfm}, JUDO~\citep{kang2026judo}, and AD-Copilot~\citep{jiang2026adcopilot}. For AUROC, we also compare against non-VLM few-shot methods: WinCLIP~\citep{jeong2023winclip}, AnomalyGPT~\citep{gu2024anomalygpt}, PromptAD~\citep{li2024promptad}, and AnomalyDINO~\citep{damm2025anomalydino}.

\noindent\textbf{Implementation Details.} Our backbone is initialized with Qwen2.5-VL-7B. CoT-SFT runs for $3$ epochs with AdamW at a learning rate of $1\times10^{-5}$. In GRPO, we sample $G=8$ responses per query with reward weights $\lambda_{\text{iou}}=0.4$, $\lambda_{\text{ans}}=0.4$, $\lambda_{\text{fmt}}=0.2$ and a dynamic length penalty. The segmentation decoder predicts at $512\times512$ and is optimized with AdamW (lr $2\times10^{-4}$) with $\lambda_{\text{bce}}=5$, $\lambda_{\text{dice}}=5$, $\lambda_{\text{const}}=2$, $\lambda_{\text{cls}}=1$, $\lambda_{\text{tv}}=1$. LoRA modules with rank $r{=}16$, $\alpha{=}32$, and dropout $0.05$ are injected into the VLM attention and MLP projections. All experiments run on $4$ NVIDIA A800 GPUs with DeepSpeed ZeRO-3.

\noindent\textbf{Model Variants and Metrics.}
All checkpoints share the Stage-1 CoT-SFT initialization $\theta_0$. Stage~2 yields InspectorGPT-Reason ($\theta_G$), Stage~3 yields InspectorGPT-Seg ($\theta_S$ with decoder $\Phi_S$), and Stage~4 fuses their task vectors into the final InspectorGPT ($\theta^*$), which uses $\alpha{=}0.3$ and is what ``InspectorGPT'' refers to throughout. The endpoints $\alpha{=}0$ and $\alpha{=}1$ recover $\theta_G$ and $\theta_S$ exactly. For VQA, we report A-Disc., the anomaly discrimination accuracy; Sem-6, the macro average of the six semantic sub-tasks; and Avg., the macro average of all seven, so that $\text{Avg.}=(\text{A-Disc.}+6\,\text{Sem-6})/7$. All ablations use the same split, protocol and four benchmarks as the main results.

\subsection{Main Results}

\noindent\textbf{Multi-dimensional Inspection on MMAD.}
Table~\ref{tab:multidimensional_results} reports the seven VQA dimensions. InspectorGPT attains the best average accuracy (82.38\%), ahead of the strongest specialist baseline AD-FM (82.03\%) and of a $10\times$ larger general-purpose VLM (Qwen2.5-VL-72B, 76.96\%), while also leading on anomaly discrimination (73.90\%) and defect analysis (88.25\%).

The table also exposes the trade-off that motivates this work. Reasoning-oriented specialists purchase semantic quality at the expense of the most elementary capability: OmniAD (68.80\%) and JUDO (64.51\%) both discriminate worse than the very base VLM they fine-tune (70.47\%), despite leading it by 2.90 and 20.08 points, respectively, on defect description. InspectorGPT is the only model in the top group on discrimination, defect analysis and the average at once, which Section~\ref{subsec:ablation} attributes to how two separately trained capabilities are assembled.

\begin{table*}[t]
\centering
{\normalsize
\setlength{\tabcolsep}{1.5mm}
\begin{tabular}{@{}lccccccccc@{}}
\toprule
\multirow{2}{*}{\textbf{Model}} & \multirow{2}{*}{\textbf{Scale}} &
\multicolumn{1}{c}{\textbf{Anomaly}} &
\multicolumn{4}{c}{\textbf{Defect}} &
\multicolumn{2}{c}{\textbf{Object}} &
\multirow{2}{*}{\textbf{Average}} \\

 & &
\textbf{Disc.} &
\textbf{Cls.} &
\textbf{Loc.} &
\textbf{Desc.} &
\textbf{Anal.} &
\textbf{Cls.} &
\textbf{Anal.} & \\
\midrule

\multicolumn{10}{@{}l}{\textbf{Proprietary}}\\
GPT-4o-mini    & -- & 64.33 & 48.58 & 38.75 & 63.68 & 80.40 & 88.56 & 79.74 & 66.29 \\
GPT-4o         & -- & 68.63 & 65.80 & 55.62 & 73.21 & 83.41 & \underline{94.98} & 82.80 & 74.92 \\
Qwen3-VL-Plus  & -- & 69.69 & 56.27 & 53.79 & 64.39 & 83.64 & 94.49 & \textbf{90.29} & 73.22 \\
Gemini 2.5 Pro & -- & 71.14 & 53.40 & 63.24 & 72.33 & 78.80 & 90.32 & 81.20 & 72.92 \\
\midrule

\multicolumn{10}{@{}l}{\textbf{Open-source}}\\
AnomalyGPT     & 7B  & 65.57 & 27.49 & 27.97 & 36.86 & 32.11 & 29.84 & 35.82 & 36.52 \\
LLaVA-OneVision & 7B  & 51.77 & 46.13 & 41.85 & 62.19 & 69.73 & 90.31 & 80.93 & 63.27 \\
InternVL2      & 8B  & 59.97 & 43.85 & 47.91 & 57.60 & 78.10 & 74.18 & 80.37 & 63.14 \\
InternVL2      & 76B & 68.25 & 54.22 & 56.66 & 66.30 & 80.47 & 86.40 & 82.92 & 70.75 \\
InternVL3.5    & 8B  & 67.50 & 49.37 & 57.90 & 58.07 & 77.66 & 72.01 & 81.11 & 66.23 \\
Qwen2.5-VL     & 7B  & 70.47 & 56.15 & 60.35 & 64.30 & 78.34 & 92.11 & 83.58 & 72.19 \\
Qwen2.5-VL     & 72B & 72.66 & 62.31 & 67.16 & 73.56 & 81.95 & 94.30 & 86.78 & 76.96 \\
\midrule

\multicolumn{10}{@{}l}{\textbf{Fine-tuned}}\\
AnomalyR1        & 3B & 60.20 & 63.50 & 70.10 & 80.40 & 85.20 & 82.40 & 86.10 & 75.41 \\
OmniAD (1-shot) & 7B & 68.80 & \underline{78.80} & 75.50 & 67.20 & 86.40 & \textbf{96.00} & 86.40 & 79.87 \\
EMIT & 8B & \underline{73.87} & \textbf{80.85} & \underline{76.39} & 83.00 & 85.92 & 90.26 & 83.37 & 81.95 \\
AD-FM & 7B & 73.15 & 73.36 & \textbf{77.53} & \textbf{86.80} & 86.75 & 89.98 & 86.67 & \underline{82.03} \\
JUDO & 7B & 64.51 & 72.17 & 75.95 & \underline{84.38} & \underline{87.76} & 94.24 & 86.07 & 80.73 \\
AD-Copilot$^{*}$ & 7B & 73.95 & 74.29 & 76.40 & 84.92 & 86.93 & 91.86 & 87.67 & 82.29 \\
InspectorGPT    & 7B & \textbf{73.90} & 75.32 & 75.91 & 82.18 & \textbf{88.25} & 92.94 & \underline{88.18} & \textbf{82.38} \\
\bottomrule
\end{tabular}
}
\caption{Comparison of proprietary, open-source, and fine-tuned models on multi-dimensional tasks over MMAD. Disc., Cls., Loc., Desc., and Anal. denote Discrimination, Classification, Localization, Description, and Analysis, respectively. All reported scores represent VQA accuracy (\%). Bold and underlined entries indicate the best and second-best results in each column, respectively. $^{*}$AD-Copilot is trained under a different data setting and is excluded from the ranking.}
\label{tab:multidimensional_results}
\end{table*}

\begin{table*}[t]
\centering
\begin{minipage}[t]{0.46\textwidth}
    \centering
    {\small
    \setlength{\tabcolsep}{1.8mm}
    \begin{tabular}{@{}lcccc@{}}
        \toprule
        \multirow{2}{*}{\textbf{Method}} & \multicolumn{2}{c}{\textbf{MVTec-AD}} & \multicolumn{2}{c}{\textbf{VisA}} \\
        \cmidrule(lr){2-3}\cmidrule(lr){4-5}
         & Image & Pixel & Image & Pixel \\
        \midrule
        WinCLIP        & 93.1 & 95.2 & 83.8 & 96.4 \\
        AnomalyGPT     & 94.1 & 95.3 & \textbf{87.4} & 96.2 \\
        PromptAD         & 94.6 & 95.9 & \underline{86.9} & \underline{96.7} \\
        AnomalyDINO      & \underline{96.6} & \underline{96.8} & \textbf{87.4} & \textbf{97.8} \\
        InspectorGPT-Seg & \textbf{96.9} & \textbf{98.7} & 83.2 & \textbf{97.8} \\
        \bottomrule
    \end{tabular}
    }
    \caption{Comparison with non-VLM few-shot methods under standard AUROC (\%).}
    \label{tab:auroc}
\end{minipage}
\hfill
\begin{minipage}[t]{0.5\textwidth}
    \centering
    {\small
    \setlength{\tabcolsep}{1.4mm}
    \begin{tabular}{@{}lccccc@{}}
        \toprule
        \textbf{Method} & \textbf{DAGM} & \textbf{DTD} & \textbf{SDD} & \textbf{MPDD} & \textbf{Avg.} \\
        \midrule
        Qwen2.5-VL & 94.83 & 85.59 & 84.17 & 68.34 & \underline{83.23} \\
        IAD-R1 & 85.20 & 83.40 & \textbf{90.80} & 65.80 & 81.30 \\
        AD-FM & \textbf{95.51} & \underline{92.64} & -- & \textbf{72.71} & -- \\
        InspectorGPT & \underline{95.08} & \textbf{93.75} & \underline{86.67} & \underline{72.12} & \textbf{86.91} \\
        \bottomrule
    \end{tabular}
    }
    \caption{Zero-shot generalization of the MMAD-trained model to unseen benchmarks (AD accuracy \%). AD-FM does not report SDD, so its average is omitted.}
    \label{tab:unseen_benchmarks}
\end{minipage}
\end{table*}

\noindent\textbf{Pixel-level Detection under AUROC.}
Generative VLMs emit deterministic text and boxes rather than continuous anomaly scores, so AUROC does not apply to VQA-style outputs. InspectorGPT-Seg produces dense masks and can therefore be placed on the standard AUROC protocol against non-VLM few-shot detectors (Table~\ref{tab:auroc}). It attains the best scores on MVTec-AD (96.9 image, 98.7 pixel) and ties the best pixel-level AUROC on VisA (97.8), confirming that adding comparative reasoning costs nothing in conventional detection quality.

\noindent\textbf{Generalization to Unseen Benchmarks.}
Had the model merely absorbed MMAD-specific response patterns, this would surface as a collapse outside MMAD. We therefore evaluate the same checkpoint on four disjoint benchmarks with no target-domain adaptation (Table~\ref{tab:unseen_benchmarks}). InspectorGPT obtains the best average (86.91\%), improving over its base model by 3.68 points and over the fine-tuned IAD-R1 by 5.61. The largest gain is on DTD (+8.16 over the base), a purely textural domain absent from training, where a fixed feature prototype is least transferable and a comparison rule most reusable.

\subsection{Ablation Study}
\label{subsec:ablation}

\noindent\textbf{Input Paradigm.}
Block~(a) of Table~\ref{tab:ablation_paradigm} compares a single-image model with our dual-image formulation. The single-image variant is trained separately under the identical recipe, so the two rows differ only in whether a normal reference is available at training and inference time. Supplying that reference raises average discrimination from 65.83\% to 73.90\% ($+8.07$), and the gain concentrates where a fixed appearance prior is least informative: $+13.49$ on VisA and $+10.16$ on GoodsAD, whose categories are visually heterogeneous. MVTec-LOCO is the exception ($-0.33$): its logical anomalies are defined by global compositional rules that a single reference image cannot express.

\noindent\textbf{Training Stages and Reasoning Strategy.}
Block~(a) of Table~\ref{tab:ablation_components} ablates the two training stages of $\theta_G$. Chain-of-Thought supervision is the load-bearing component: retraining the identical pipeline on direct-answer data collapses A-Disc.\ to 52.66\%, barely above the chance level of a binary decision, because without an explicit comparison trace the model has no mechanism for relating the two images and falls back on single-image appearance matching. The two optimization stages are complementary rather than redundant: SFT alone reaches 73.68\% and GRPO applied to the base model 73.45\%, yet their composition reaches 82.50\%. SFT installs the parsable comparative format to which GRPO can subsequently assign credit; neither is sufficient in isolation.

The block also reproduces the discrimination collapse discussed in the introduction under controlled conditions: the base VLM discriminates at 70.47\%, CoT-SFT drops it to 67.48\%, and GRPO recovers it only to 69.57\%, still below the untuned starting point. Reasoning training thus buys 12.19 points of Sem-6 at the expense of the binary decision itself. Because every row shares one architecture, split and protocol, this isolates the reasoning objective as the cause rather than any particular published recipe.

\noindent\textbf{Reward Design in GRPO.}
Block~(b) of Table~\ref{tab:ablation_components} removes one reward term at a time. $R_{\text{answer}}$ is the most influential ($-10.60$ average), expected since it is the only term that scores task correctness; removing it also costs the most on the semantic dimensions ($-12.21$ Sem-6). $R_{\text{iou}}$ costs 7.56 points, concentrated on the grounding-dependent dimensions. $R_{\text{format}}$ never scores correctness at all, yet deleting it costs 8.30 points of Sem-6, nearly as much as $R_{\text{iou}}$ does (8.74): structural parsability is a precondition for the other two rewards to be assigned, and an unparsable trajectory receives no learning signal regardless of whether its content is correct.

\begin{table}[!htb]
    \centering
    {\small
    \setlength{\tabcolsep}{1mm}
    \begin{tabularx}{\columnwidth}{@{}Xccc@{}}
        \toprule
        \textbf{Variant} & \textbf{A-Disc.} & \textbf{Sem-6} & \textbf{Avg.} \\
        \midrule
        \multicolumn{4}{@{}l}{\emph{(a) Training stages \& reasoning}}\\
        Base (Qwen2.5-VL-7B)   & 70.47 & 72.47 & 72.19 \\
        w/o SFT                & 67.82 & 74.39 & 73.45 \\
        w/o CoT                & 52.66 & 76.06 & 72.72 \\
        w/o GRPO               & 67.48 & 74.71 & 73.68 \\
        \midrule
        \multicolumn{4}{@{}l}{\emph{(b) Reward design in GRPO}}\\
        w/o $R_{\text{answer}}$ & 68.56 & 72.45 & 71.90 \\
        w/o $R_{\text{iou}}$    & 69.04 & 75.92 & 74.94 \\
        w/o $R_{\text{format}}$ & 71.24 & 76.36 & 75.63 \\
        \midrule
        \multicolumn{4}{@{}l}{\emph{(c) Branch assembly \& fusion}}\\
        InspectorGPT-Reason ($\theta_G$, $\alpha{=}0$) & 69.57 & \textbf{84.66} & \textbf{82.50} \\
        InspectorGPT-Seg ($\theta_S$, $\alpha{=}1$)    & \textbf{79.08} & 74.31 & 74.99 \\
        Jointly trained                                & 69.08 & 81.67 & 79.87 \\
        InspectorGPT ($\theta^*$, $\alpha{=}0.3$)      & \underline{73.90} & \underline{83.80} & \underline{82.38} \\
        \bottomrule
    \end{tabularx}
    }
    \caption{Unified ablation on MMAD (VQA accuracy \%). Rows in (a) and (b) are ablations of InspectorGPT-Reason. Bold and underlined entries indicate the best and second-best results within block (c), respectively.}
    \label{tab:ablation_components}
\end{table}

\begin{table}[!htb]
    \centering
    {\small
    \setlength{\tabcolsep}{0.8mm}
    \begin{tabular}{@{}lccccc@{}}
        \toprule
        \textbf{Setting} & \textbf{MVTec-AD} & \textbf{VisA} & \textbf{LOCO} & \textbf{GoodsAD} & \textbf{Avg.} \\
        \midrule
        \multicolumn{6}{@{}l}{\emph{(a) Input paradigm --- InspectorGPT}}\\
        Single-image & 77.30 & 65.66 & \textbf{64.19} & 56.15 & 65.83 \\
        Dual-image   & \textbf{86.29} & \textbf{79.15} & 63.86 & \textbf{66.31} & \textbf{73.90} \\
        \midrule
        \multicolumn{6}{@{}l}{\emph{(b) Seg-LoRA --- InspectorGPT-Seg}}\\
        w/o LoRA & 89.93 & 64.18 & 56.87 & 58.92 & 67.48 \\
        w/ LoRA  & \textbf{90.28} & \textbf{83.70} & \textbf{67.11} & \textbf{75.24} & \textbf{79.08} \\
        \bottomrule
    \end{tabular}
    }
    \caption{Ablation on the input paradigm and on the Seg-LoRA injection. Anomaly discrimination accuracy (\%) is reported in both blocks. Bold entries indicate the best result in each block and column.}
    \label{tab:ablation_paradigm}
\end{table}

\noindent\textbf{Seg-LoRA Injection.}
Block~(b) of Table~\ref{tab:ablation_paradigm} isolates the LoRA co-training stage. When the decoder is trained on a frozen backbone, the VLM still answers with the Stage-1 weights $\theta_0$ and its discrimination remains at 67.48\%, identical to the \emph{w/o} GRPO row above; the pixel-level objective improves the mask but never reaches the text head. Opening that gradient path through Seg-LoRA raises average discrimination to 79.08\% ($+11.60$), and the gains track the density of the available evidence: $+16.32$ on GoodsAD and $+19.52$ on VisA. MVTec-LOCO improves as well ($+10.24$), reversing the loss observed in the input-paradigm ablation, because dense masks supply the structural signal about part layout that a single reference image could not convey.

\begin{figure}[t]
    \centering
    \includegraphics[width=\columnwidth]{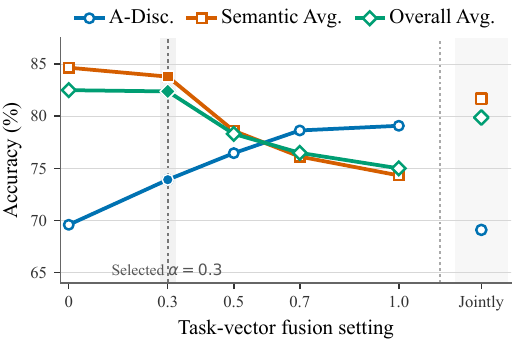}
    \caption{Ablation of the task-vector fusion coefficient $\alpha$. We select $\alpha{=}0.3$, which buys 4.33 points of A-Disc.\ for 0.12 points of overall average.}
    \label{fig:fusion_ablation}
\end{figure}

\noindent\textbf{Task-Vector Fusion.}
Block~(c) of Table~\ref{tab:ablation_components} compares the two branches with the assembled model. They are unbalanced in opposite directions: $\theta_G$ is the stronger semantic model (Sem-6 84.66\%, A-Disc.\ 69.57\%) and $\theta_S$ the stronger discriminator (A-Disc.\ 79.08\%, Sem-6 74.31\%). Optimizing both objectives on a single model does not reconcile them: joint training reaches only 79.87\% average and leaves A-Disc.\ at 69.08\%, forfeiting 2.99 points of Sem-6 relative to $\theta_G$ without buying discrimination in return. Fusing the two task vectors, which updates no weight, improves on it by 4.82 points of A-Disc.\ and 2.51 of average accuracy.

Figure~\ref{fig:fusion_ablation} sweeps the coefficient $\alpha$. A-Disc.\ rises monotonically from 69.57\% to 79.08\% while Sem-6 falls from 84.66\% to 74.31\%, but the exchange rate is far from constant: the first 0.3 of the sweep delivers 4.33 points of A-Disc.\ for 0.86 of Sem-6, whereas the remaining 0.7 costs 9.49 points of Sem-6 for 5.18. The asymmetry suggests that the two task vectors are largely non-interfering near the reasoning endpoint, and begin to compete only once the segmentation direction dominates. We therefore set $\alpha{=}0.3$, which lifts discrimination above the untuned base VLM for the first time (73.90\% vs.\ 70.47\%) at a cost of 0.12 points of overall average.

%% file: content/conclusion.tex
\section{Conclusion}
We present InspectorGPT, an interpretable and generalizable vision-language framework for industrial anomaly detection. A staged pipeline couples large-model semantic reasoning with precise localization, while our GRPO reward engine suppresses unfounded diagnoses through verifiable answer--box consistency. A segmentation decoder, LoRA co-training, and task-vector fusion balance semantic generation and anomaly discrimination. Although task-vector fusion provides an initial balance, further exploration is needed.